# Developing an AI Course for Synthetic Chemistry Students

Zhiling Zheng

Department of Chemistry, Washington University, St. Louis, Missouri 63130, United States

**ABSTRACT:** Artificial intelligence (AI) and data science are transforming chemical research, yet few formal courses are tailored to synthetic and experimental chemists, who often face steep entry barriers due to limited coding experience and lack of chemistry-specific examples. We present the design and implementation of AI4CHEM, an introductory data-driven chemistry course created for students on the synthetic chemistry track with no prior programming background. The curriculum emphasizes chemical context over abstract algorithms, using an accessible web-based platform to ensure zero-install machine learning (ML) workflow development practice and in-class active learning. Assessment combines code-guided homework, literature-based mini-reviews, and collaborative projects in which students build AI-assisted workflows for real experimental problems. Learning gains include increased confidence with Python, molecular property prediction, reaction optimization, and data mining, and improved skills in evaluating AI tools in chemistry. All course materials are openly available, offering a discipline-specific, beginner-accessible framework for integrating AI into synthetic chemistry training.

## INTRODUCTION

How prepared are experimental chemists to work within an AI driven research landscape? Nowadays machine learning and data-driven methods influence many aspects of modern synthetic chemistry, from small-molecule property prediction[1-4] and materials design[5-7] to reaction optimization[8-12] and automated synthesis[13-17]. These approaches increasingly shape how new synthetic methods are developed and how experimental campaigns are planned.[18-22] In practice, however, most chemistry curricula offer little formal training in machine learning or, more broadly, artificial intelligence.[23-25] This gap leaves both undergraduate and graduate students on synthetic tracks whose primary identity is that of a bench chemist underprepared to leverage AI in their laboratory work.

For experimental chemists in particular, two barriers consistently appear. First, students who have spent most of their training on synthesis, reaction mechanisms, or instrumentation often have little or no exposure to Python programming and advanced mathematics.[25-28] At the same time, many ML textbooks and online resources rely on abstract examples or datasets (e.g. from finance or image analysis), which may feel distant from the challenges encountered in synthetic chemistry. When instruction lacks chemical context, chemistry-major learners struggle to see how the course connects to their experimental practice and may view AI as overly theoretical.[29-31] There is a clear need for beginner-friendly instruction that grounds AI concepts in chemical scenarios,[26,32] helping students see direct links to wet-lab tasks and scientific questions they care about.

To meet these needs, effective AI education for this experimental chemists audience should meet three goals: (i) assume minimal programming background and present tools in a staged manner that keeps cognitive load manageable, (ii) use chemical datasets and reaction scenarios so that students can connect AI/ML concepts to ideas such as functional groups, spectroscopic patterns, and reaction yields, and (iii) cover the complete workflow around an experiment, including data collection and cleaning, model choice and evaluation, visualization, and planning of subsequent reactions.

In response to these needs, we developed *AI for Experimental Chemistry* (AI4CHEM), an introductory curriculum built around real-world chemical systems and tasks that experimental chemists routinely encounter, such as property prediction, reaction optimization, and the use of intelligent agents to support laboratory analysis. Through this discipline-aligned framing, AI4CHEM seeks to lower the entry barrier for chemistry majors and graduate students and to cultivate the foundational skills needed to incorporate data-driven thinking into daily research practice.

## COURSE DESCRIPTION AND RATIONALE

The goal of this one-semester course is to provide a structured and approachable introduction to the application of AI in chemistry (Figure 1) for upper-level undergraduates and graduate students beginning research in synthetic organic, inorganic, and materials chemistry as well as biochemistry. The course serves as an entry point for bench chemists who want to understand how AI tools can complement experimental work but who may not have prior exposure to programming and data science. The design centers on using chemical datasets (e.g., molecular structures, reaction conditions, spectral measurements, microscopic images, and laboratory records) as the primary medium through which ML concepts are introduced. Students practice building workflows that reflect the decisions chemists make in the laboratory, such as correlating structure with reactivity, identifying trends across families of substrates, and optimizing reactions based on previous trial-and-error.

All instructional materials use real chemical examples and experimental datasets so that students immediately recognize the relevance of each concept to their work in synthesis and characterization.

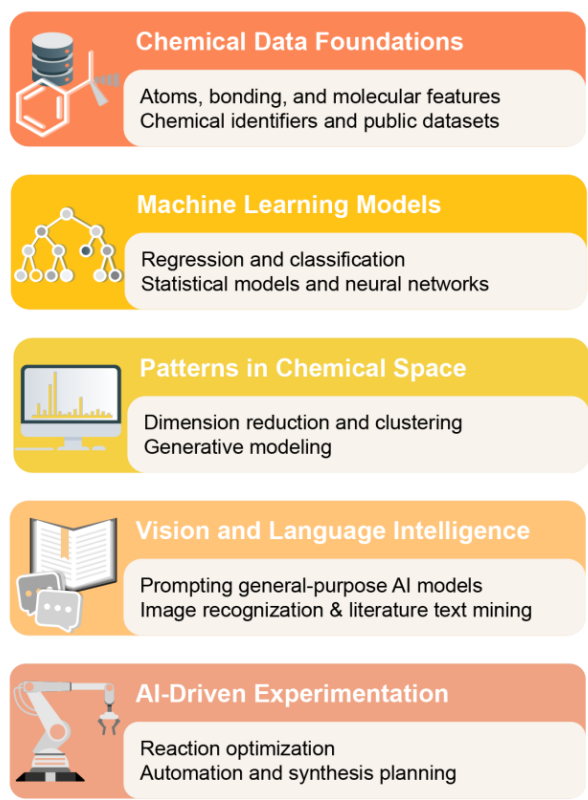

**Figure 1.** Conceptual design of the AI4CHEM curriculum organized around five themes in AI for experimental chemistry. The modules guide students from representing and visualizing realistic chemical datasets to building and interpreting predictive models, and using AI tools to plan experiments. This progression is intended to help synthetic chemists connect data-driven methods to tasks they routinely face in the wet lab.

The course outcomes were formulated to reflect both AI and chemistry disciplinary goals (Table 1). By the end of the semester, students are expected to be able to

(i) Communicate AI and ML concepts in chemistry clearly in written and oral form.
(ii) Identify and select appropriate machine learning approaches for a range of chemical data analysis and prediction tasks.
(iii) Interpret common cheminformatics workflows and apply ready-to-run code to support experimental planning.
(iv) Use regression and classification models to predict molecular properties and reaction outcomes.
(v) Analyze microscopy and imaging data with basic computer vision techniques.
(vi) Employ generative and language model tools to support molecular design, experiment planning, and literature study.

Several pedagogical goals were considered during development of the course. First, we sought to lower the barrier of entry for students without a coding background by integrating Python instruction directly with chemical examples rather than treating programming as a separate prerequisite. Second, in line with prior work that emphasizes the importance of discipline-specific contexts for programming in chemistry,[25,26,30,32] every topic is anchored in a chemical scenario that students might realistically encounter in their own research. Third, we adopted an active learning model in which students spend significant class time working through online notebook-based tutorials in small groups, with the instructor circulating to provide just-in-time support. Finally, we aimed to cultivate critical thinking about AI through repeated attention to data quality, model limitations, and ethical considerations connected to AI use in chemistry.[24,33–35]

The course meets twice weekly for 80 minutes for a total of approximately 30 class periods. Each class session is structured with roughly 40 minutes devoted to interactive lecture and whole-class discussion and 40 minutes to hands-on tutorial work. During lectures, we introduced concepts and highlighted examples, while the tutorial portion allowed students to directly apply those ideas by working through notebook exercises with guidance. To support interaction during coding activities, enrollment is capped at 15 students; the inaugural offering enrolled four advanced undergraduates and nine first-year graduate students. Most students had backgrounds in different subfields of chemistry (organic, inorganic, materials, etc.) and enrolled to enhance their research with data-driven methods. This diversity informed our content design as we assumed only general chemistry and introductory organic chemistry knowledge and we started from first principles in both coding and machine learning. Benefit from the small class size, the primary teaching methods thus blended short lectures, live demonstrations, guided discussions connected to recent literature, and collaborative coding practice work in which students implement and modify AI workflows, enabling one-on-one assistance during the hands-on activities and fostering an inclusive environment for students who were uncertain about programming.

An important aspect of our course design was the inclusion of current research examples to inspire and engage students.[26,36] We incorporated a few cutting-edge applications of AI in chemistry as case studies (see Section COURSE CONTENTS), not for detailed technical instruction but to illustrate the state-of-the-art and various AI applications. Our intent was to show students why learning these techniques matters by connecting foundational concepts to real breakthroughs. In addition, we are aware that linking basic methods to significant advances provides context and motivation for students to apply ML in their own research. Indeed, exposing students to how AI is used in modern chemistry – from optimizing metal–organic framework syntheses to analyzing big data from catalysis – helped them envision applying similar approaches in their graduate projects. Overall, the course rationale was to provide a discipline-specific, beginner-accessible framework that empowers experimental chemists to engage with AI, thereby addressing a critical skills gap in chemistry education.

**Table 1. Weekly organization of the AI4CHEM curriculum with each module aligns core AI concepts with corresponding chemical concepts and in class tutorials.**

| Module (Number of Weeks) | Machine Learning Principles | Chemistry Principles | In Class Tutorials |
| --- | --- | --- | --- |
| Python Foundations & Chemical Data (1) | Data types, lists/arrays, plotting, visualization. | Units, concentrations, calibration curves. | • Plot Beer–Lambert calibration curve<br>• Calculate molarity from mass and volume |
| Molecular Representation & Cheminformatics (1) | SMILES notation, atom/bond features, canonicalization, molecular fingerprints. | Functional groups, valence, aromaticity, stereochemistry, charged fragments, motifs. | • Draw molecules with RDKit<br>• Compute similarity between substrates |
| Classical ML for Molecular Properties (2) | Supervised learning, regression, classification, data splitting, cross-validation, feature importance. | Property trends: melting point, solubility, polarity, structure–activity relationships, heterogeneous experimental data. | • Predict melting point from descriptors<br>• Classify solubility categories and compare toxicity models |
| Neural Networks (2) | Neural layers, activation, graph message passing, model explanation. | Bond energies, electron distribution, substituent effects, steric/electronic environments around reactive centers. | • Build MLP to predict molecular logS values<br>• Predict substituent effects for C-H oxidation reactions. |
| Self-Supervised Embeddings and Generative Modeling (1) | PCA, t-SNE, UMAP, clustering., encoder, decoder, latent space. | Scaffold families, reactivity groupings, structure–function patterns, synthesis feasibility. | • Cluster electrochemical reaction substrates using learned embeddings<br>• Generate drug-like molecules from VAE latents |
| Reaction Optimization & Closed-Loop Design (1) | Gaussian processes, acquisition functions, Bayesian optimization, active learning, reinforcement learning policy. | Reaction yield surfaces, catalyst and solvent screening, balancing yield vs selectivity, iterative experiment selection. | • Optimize yield with BO surrogate<br>• Synthesize metal-organic frameworks with multi-objective optimization<br>• Screen catalysts via active learning |
| Multimodal Chemical Intelligence (1) | Convolutional neural network, CLIP, multimodal retrieval, transformer architecture. | Microstructure features, spectral patterns, qualitative materials analysis via visual signatures. | • Retrieve data with CLIP image–text pairs<br>• Classify crystal images with CNN and VLM.<br>• Build a tiny transformer to predict molecular reactivities. |
| Literature Data Mining & Lab Automation (2) | Prompt engineering, name entity recognition, JSON schema outputs, chain-of-thought reasoning, structured outputs, agentic workflows, instrument APIs. | Identifying reaction types, extracting catalysts/solvents, evaluating reproducibility, minimizing waste, sustainable experiment design. | • Extract reaction conditions from literature using prompts<br>• Build JSON synthesis tables<br>• Plan closed-loop experiment with agentic workflow |

In addition, a key consideration in our course implementation was how to conduct the hands-on cheminformatic components in a way that minimized technical barriers and maximized accessibility for all students. From the outset, AI4CHEM was conceived as a "dry-lab" course in which students carry out computational experiments over authentic chemical datasets that closely resemble those encountered in experimental research. To make these experiences accessible to students with a wide range of computational backgrounds, in practice, all instruction is delivered through interactive web platform (a Jupyter Book, Figure 2a) and executed via Google Colab cloud notebooks (Figure 2b). This setup meant that any student could participate using only a web browser, with no need to configure their own computer for programming. Google Colab provided a free, cloud-based Python environment with all necessary libraries pre-installed, which dramatically lowered the entry barrier for novices. By clicking a Colab link, students could immediately run code and see results, whether on a personal laptop or a university computer cluster. We found this approach crucial for inclusivity and it leveled the playing field between students who had prior coding experience and those who had never opened a terminal. Furthermore, we made the entire course website publicly accessible (no login required), anticipating that our materials could be used as an open educational resource by other chemistry students or educators outside our class.

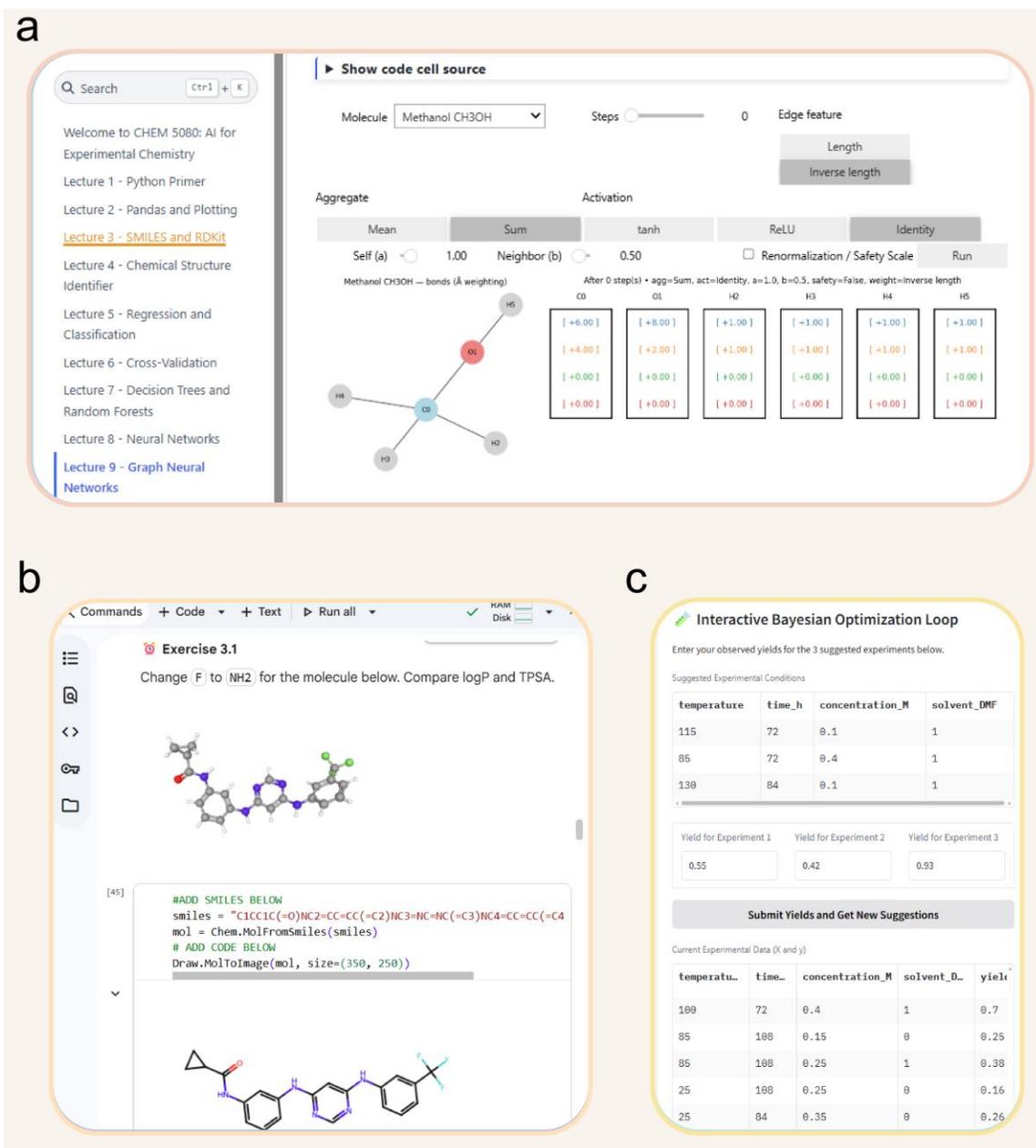

**Figure 2.** Example interactive learning interfaces used in AI4CHEM. (**a**) Public course website that organizes lecture notes and tutorials and allows students to explore interactive visualizations (e.g., demo of graph neural networks for molecular structures). (**b**) Google Colab tutorial in which students run and modify Python code with no pre-installation needed to edit functional groups and compare computed molecular properties. (**c**) Undergraduate student developed interactive Bayesian optimization loop user interface created as a homework assignment, where chemists can enter observed reaction yields, view suggested conditions, and obtain updated recommendations for the next experiments.

The Jupyter Book presents each module as an HTML page with static web rendering that allows students to peruse the material on any device (laptop, tablet, even smartphone) without needing a Python environment. However, whenever they are ready to dive deeper, a prominent button on each page opens the same notebook in Google Colab for live interaction. This website served as a central hub for all lecture notes and tutorials in an organized, interactive format. Within a module, short explanatory text cells alternate with code cells, visualizations, and interactive widgets. Our previous experience found that the ability to toggle between a formatted textbook view and an executable notebook proved very effective. It let students learn at their own pace. For example, those less comfortable with coding could stick to the guided structure, whereas advanced students could modify code and experiment further. Importantly, this open notebook structure also allowed students to revisit any lesson after class and even share the materials with peers, since everything remained accessible online.

On the other hand, the Colab tutorials are interactive "dry-lab" tutorials where students actively worked through programming tasks with instructor support. We treated these sessions much like a digital chemistry laboratory where students were encouraged to explore, make mistakes, ask questions, and learn by doing. To scaffold the learning process, every notebook includes extensive comments, chemical context, and multiple visualization checkpoints so that readers can see intermediate results without reading large blocks of code. In-line exercises prompt students to change parameters, extend plots, or interpret model outputs, and many notebooks include small "mini-games", such as navigating the reaction yield surface or predicting which catalyst will be most reactive. By interleaving code execution with immediate visual feedback and discussion, the notebooks keep students engaged and help demystify what the algorithms are doing. We found that this approach fostered a highly interactive classroom atmosphere since often a student would run a code cell and then ask, "*Why did this molecule's prediction error turn out large?*" or "*What happens if we remove this outlier point?*" Such moments opened up rich discussions linking back to chemistry (e.g., examining if that molecule was an outlier due to an unusual functional group). The instructor circulated to assist with errors, but over time students became more confident in debugging simple problems and even helping each other.

Several practical considerations were important in making the computer labs effective for beginners. First, we included detailed comments and guidance in the code cells. For instance, if a code block was importing a library or performing a tricky step, we prefaced it with human-readable explanations (in comments or markdown) so that students understood why each step was happening. Long code snippets were broken into smaller chunks with intermediate outputs, to avoid overwhelming novice programmers with walls of code. Second, we intentionally used consistent coding patterns throughout the semester. Early on, students learned a template for tasks like training a model or plotting data, and we reused these patterns in later notebooks. This repetition built familiarity and reduced cognitive load when new concepts were introduced so that students could focus on the concept rather than getting lost in new syntax. Third, in preparation for class, we often assigned students to preview the notebook or read a short article ahead of time. While not all students fully completed pre-reading, even a cursory glance meant they came in with some context, making the live session more productive.

Another innovative element of our course was encouraging students to treat the notebooks as playgrounds for experimentation. We explicitly told students that after completing the given tasks, they should feel free to change inputs, test hypotheses, or even break things to see what happens, which is much as a chemist might tinker with reaction conditions in a wet-lab. This invitation to explore led to some creative learning moments. For example, in one exercise involving a neural network predicting molecular toxicity, a student wondered how the predictions would change if a particular substructure was removed; they edited the code to alter the molecule and re-ran it, sparking a whole discussion on feature importance and model extrapolation. By the end of the course, students were comfortable enough to even develop small independent projects within the notebook framework. Notably, one undergraduate in the class built a mini interactive app for the assignment: it allowed users to input reaction conditions and then used a trained Bayesian model to suggest the next experiment (Figure 2c). This was a remarkable achievement for a student who started the semester with no coding background and demonstrated the efficacy of the in-class tutorial-driven approach in empowering learners. It also resulted in a tangible tool that the student's research lab could potentially use for planning experiments.

Finally, the decision to make all course materials open access is worth highlighting. All notebooks, data files, and the course website are openly available on the course webpage and the associated GitHub page (see Supporting Information). We believe this enables more chemistry students to learn AI. Already, several students in our institution not enrolled in the class have accessed the HTML notebooks and reported learning new skills on their own. By sharing our Jupyter Book publicly, we enable any chemistry educator to adopt or adapt these materials for their own courses, and any student to self-train in AI with a chemistry emphasis. We belie that this aligns with a growing movement toward open educational resources in science and has the practical benefit that the content can be continuously improved by the community.

## COURSE CONTENTS

### Chemical Data Foundations.

At the beginning of the semester, the first lecture provides an orientation to contemporary AI applications in chemistry. Students explore short, instructor-guided examples of retrosynthesis prediction,[37] molecular property prediction,[1] protein structure prediction,[38] synthesizability evaluation,[39] and automated data mining,[40] using online website interface, published figures and short video demonstrations. These examples function as a conceptual map that helps students connect the five course themes to real research questions before they encounter the underlying algorithms in detail and start to appreciate that all the discoveries are built on coding and connection towards chemical-data-centered development.

Following on this, the first thematic block of tutorials, *Chemical Data Foundations*, introduces Python and basic data handling in the context of familiar general and physical chemistry topics. Students begin by using Python as a calculator to compute molar masses, convert between mass and moles, and evaluate concentration expressions for simple solution problems. They then learn to use the pandas library to load Beer–Lambert calibration curves from comma-separated value files, clean and filter tabulated data, calculate summary statistics, and create line and scatter plots of absorbance versus concentration. This approach allows students to associate new programming constructs such as lists, dictionaries, and plotting commands with chemical quantities they already understand.

Subsequent tutorials in this block introduce cheminformatics concepts through SMILES strings[41] and the RDKit and Scikit-learn library.[42] Students use RDKit to parse SMILES, draw molecules, compute basic descriptors such as molecular weight and logP, and make simple structure edits such as substituting a halogen or adding a methyl group. They also query public resources such as PubChem[43] and the NCI Chemical Identifier Resolver to convert between names, registry numbers, and machine-readable identifiers. These exercises establish a concrete pipeline for students by starting from a list of reagents or substrates so that learners can construct a chemically meaningful dataset that includes structures, identifiers, and numerical features ready for use in later machine learning models, showing them such workflow can save their time during laboratory inventory search.

### Machine Learning Models in Chemical Context

The second thematic block focuses on supervised machine learning models framed explicitly around questions that experimental chemists ask. Tutorials introduce regression and classification through small datasets where the target variables are melting points, solubility classes, toxicity labels, or reaction yields. Students construct train, validation, and test splits, fit linear and logistic regression models, and interpret performance metrics such as mean absolute error, $R^2$, accuracy, and ROC-AUC. Throughout, the instructor emphasizes connections between these metrics and experimental concepts such as measurement error, reproducibility, and predictive uncertainty.

As the semester progresses, students explore more flexible models such as decision trees, random forests, and multilayer perceptron neural networks, followed by introductory graph neural networks implemented in PyTorch.[44] All tutorials adopt a "chemistry-first" style where code cells are prepopulated with a working baseline model, and students are guided to inspect how specific changes, such as including an additional descriptor or altering the architecture of a neural network, affect predictions on a chemical task. For example, in one unit students train a Chemprop-style message-passing model to predict C–H oxidation reactivity from substrate structures,[45] then use feature importance scores and atom-level attributions to relate model outputs to known directing groups and steric environments. These experiences help students see and appreciate that machine learning models as extensions of structure–property reasoning rather than opaque prediction engines.

### Patterns in Chemical Space

The third theme, Patterns in Chemical Space, introduces unsupervised learning methods that help students visualize and organize large collections of molecules and reactions. Tutorials begin with dimensionality reduction techniques such as principal component analysis, t-distributed stochastic neighbor embedding, and UMAP. Students apply these methods to descriptor sets and fingerprint vectors, creating two-dimensional maps that reveal clusters of substrates, ligands, or metal-organic framework (MOF) building blocks. In discussion, they interpret these maps in terms of functional groups, scaffold families, or regions of reactivity, and consider how such views could inform library design or substrate selection.

Later activities extend from visualization to clustering and generative modeling. Students implement K-means and hierarchical clustering workflows that group electrochemical substrates or reaction conditions based on similarity patterns, then evaluate different choices of cluster number using elbow and silhouette criteria. In a capstone tutorial for this block, students interact with a simple variational autoencoder trained on a set of drug-like molecules.[46] By sampling and interpolating in latent space, students can observe generative models propose new SMILES strings, visualize the corresponding structures, and discuss synthetic feasibility and chemical diversity.

### Vision and Language Intelligence

The fourth theme, Vision and Language Intelligence, connects computer vision and large language models to experimental chemistry. In an initial computer vision tutorial, students build a compact convolutional neural network to classify simple image datasets[47] and then transfer those skills to chemical images, such as microscope or crystal photographs. They examine misclassified examples and discuss how visual cues such as crystal habit, color, or texture relate to underlying physical properties.[48]

Subsequent tutorials introduce multimodal models and transformer-based language models.[49-51] Using precomputed embeddings from a CLIP-style model,[52,53] students perform zero-shot classification of simple crystal or materials images and retrieve images that match short text prompts describing microstructure or apparent morphology.[54] On the language side, students design prompts and few-shot examples that guide a large language model to extract reaction conditions, yields, and catalysts from literature abstracts and to output the information in structured JSON form.[40,55] Students can compare model-extracted data to the source text, identify hallucinations or misinterpretations, and discuss the importance of domain constraints and human oversight when using AI systems to interpret chemical information.

### AI-Driven Experimentation

The final thematic block centers on AI-driven experimentation and closed-loop workflows. Tutorials introduce Bayesian optimization in the context of reaction condition screening. Students work with a toy Suzuki–Miyaura coupling dataset and a MOF synthesis dataset, fitting Gaussian process surrogate models over temperature, time, and concentration, then using acquisition functions to propose new conditions that balance exploration of uncertain regions with exploitation of promising areas.[56] Interactive plots show in real time how the surrogate model and acquisition surface update as new data are added, helping students build intuition for the exploration–exploitation tradeoff.

Building on this foundation, students explore extensions to multi-objective optimization,[9] reinforcement learning,[57,58] and positive-unlabeled learning[59] using simulated data from catalyst screening[58] and green synthesis[60,61]. For example, students learned to code to develop interface which can enter experimental yields for a series of suggested reactions and observe how a Bayesian optimization loop proposes the next set of conditions.[10,45] In another, they frame an experimental design problem as a bandit task and compare different exploration strategies.[58] The block concludes with a high-level discussion of self-driving laboratories and multi-agentic systems in which students map out the automation platforms, models, and decision rules that would be required to close the loop between AI-generated suggestions and automated experimentation in their own research areas.

At a fundamental level, these tutorial blocks operationalize the five course themes and provide a coherent sequence of "dry-lab" experiences that connect AI methods directly to chemistry practice, allowing students to see how modern AI methods interact with wet-lab workflows and how algorithmic suggestions should be interpreted alongside experimental judgment. Because the tutorials are built on open, browser-based tools and use datasets and examples drawn from synthetic and experimental chemistry, they can be reused or adapted by instructors who wish to introduce AI concepts to chemists with little prior programming experience.

## STUDENT ASSESSMENT

Assessment in AI4CHEM combined short quizzes, programming-based homework, a written mini review, and a final group project. Overall, student performance on the assessments aligned well with the intended course outcomes (Table 2). In particular, short in-class quizzes focused on conceptual understanding and asked students to interpret model behavior, reflect on workflow diagrams, or evaluate AI use cases in the literature. These brief checks allowed the instructor to monitor comprehension without requiring students to write code under time pressure.

**Table 2. Percentage student achievement of course outcome goals**

| Course Outcome Goal | Percentage Achieved |
|---|---|
| Communicate chemical and ML ideas clearly in written and oral form | 13 of 13 (100%) |
| Identify suitable ML approaches for chemical data analysis and prediction tasks | 12 of 13 (92%) |
| Interpret cheminformatics workflows and apply ready-to-run code for planning and analysis | 10 of 13 (77%) |
| Apply classification and regression methods to predict molecular properties and reaction outcomes | 9 of 13 (69%) |
| Analyze microscopy and imaging data using computer vision techniques and connect results to scientific problems | 11 of 13 (85%) |
| Use generative and language model tools to support molecular design, experiment planning, and literature study | 12 of 13 (92%) |

Homework assignments were delivered as Google Colab notebooks and structured to build independence gradually. The first three assignments contained ten programming questions each, beginning with partially completed code and progressing toward full, student-written solutions. The fourth and fifth assignments functioned as mini projects in which students applied course workflows to new datasets. Because many chemical problems admit multiple correct programming solutions, assignments specified the expected printed output or plot, and any solution that reproduced those results earned full credit.

Midway through the semester, students completed a 2–4 page ACS-style mini review describing how AI or machine learning could support research in their subfield. This assignment served two goals: (i) strengthening students' ability to communicate chemical ideas involving AI and (ii) prompting them to reflect on opportunities and limitations of AI in realistic laboratory settings. Each paper was reviewed by a classmate using a structured rubric and an AI-generated-text detection tool. Peer reviewers were asked to comment only on flagged sections where hallucinations appear (e.g. fabricated reference, incorrect numbers and facts, etc.). No penalty was given if some content can likely be AI-generated and this activity was mainly designed to build awareness of hallucinated content and foster ethical writing practices. Students reported that seeing how their peers approached similar topics helped them reflect on their own writing practices and on responsible use of AI tools in scientific communication.[33,62]

The final assessment was a group project in which teams designed and demonstrated an AI-assisted workflow for a chemistry problem of their choosing. Projects included tasks such as optimization of reaction conditions, clustering microscopy images, identifying important sequences, or literature data extraction. Each group submitted a short executive summary, shared a live demonstration or mock interface using real or literature-derived data, and delivered a final presentation. The presentation was guided to contain (i) a clearly articulated chemical question or challenge, (ii) at least one code snippet demonstrating a working AI or ML component applied to chemical data, and (iii) a proposed workflow that situated the computational component within a realistic experimental context. Importantly, evaluation weighed technical accuracy, clarity of narrative, and the appropriateness of the workflow for the stated chemistry problem. The presentation component allowed students to articulate the rationale behind their design choices and respond to questions from peers and the instructor. It was found that this final project produced creative solutions of using AI and demonstrated that students were able to integrate skills acquired across the semester.

## EVALUATION AND EXTENSION OF THE COURSE

Formal and informal evaluations indicated that AI4CHEM effectively supported students' learning and shifted their attitudes toward the role of AI in experimental chemistry. Survey data (Figure 3) revealed strong learner endorsement of the tutorial-centered approach. More than half of respondents identified the in-class Colab tutorials as the component that most helped them learn AI concepts, followed by group projects and homework assignments. This underscores the importance of active, hands-on engagement for novices in AI-driven experimental chemistry.

On the other hand, students' self-reported likelihood of applying AI or ML in future research increased substantially from the beginning to the end of the course. Before the semester, only one student described themselves as "very likely" to use AI tools in their research; after completing the course, eight selected "very likely" and four selected "somewhat likely," illustrating a notable shift in confidence and perceived relevance through the better understanding of the AI concepts and examples of their applications in chemistry.

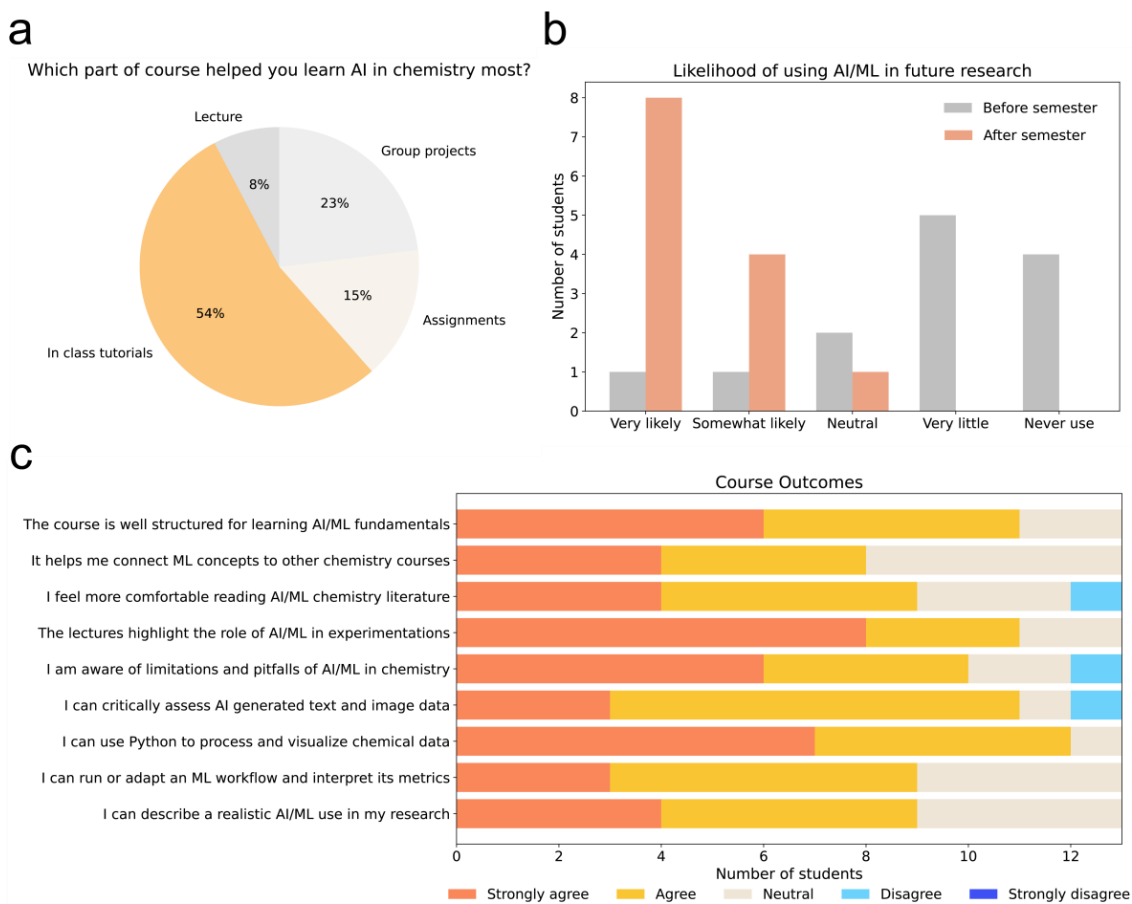

**Figure 3.** Graphical representation of learner (N = 13) responses end of course survey. (**a**) Pie chart showing which course components students felt helped them learn AI in chemistry most. (**b**) Self-reported likelihood of using AI or machine learning in future research before and after the course. (**c**) Stacked bar chart of agreement with course outcome statements.

Agreement with course outcome statements was also high. Most students reported increased comfort with reading AI-rich literature, interpreting chemical data in Python, and evaluating AI-generated content. Gains were especially pronounced in understanding model limitations, identifying common pitfalls in data-driven chemistry, and linking ML concepts to experimental design. Several students also highlighted the real-world examples and the open-access course materials as valuable features, noting that they intended to revisit the Jupyter Book after the semester to adapt workflows for research problems.

In terms of future course extension, feedback from the inaugural offering suggests several opportunities for continued refinement. First, while students were able to complete the more advanced modeling tasks, many expressed interest in additional optional practice sets focused on debugging, data cleaning, and model interpretation. Second, students requested more examples of chemical imaging and spectroscopy data, reflecting strong interest in multimodal applications.

Because all teaching materials, datasets, and example solutions are hosted publicly through a website and associated GitHub repository (see Supporting Information), the course can be adapted at other institutions in several formats. Instructors may adopt AI4CHEM as a full-semester elective, integrate selected modules into existing physical or organic chemistry courses, or use the notebooks in the context of short summer schools or research group workshops. As discussed earlier, class size was intentionally small to maintain high instructor–student interaction during tutorials; however, the instructional model could scale if supported by teaching assistants or peer mentors.

In addition, within the home institution, the AI4CHEM materials have already been used informally by students from both College of Arts & Sciences and School of Engineering who were not enrolled in the course but who access the open website for self-study.

In the next iteration of this course, we plan to incorporate short pre-class coding warm-ups and optional challenge notebooks for advanced learners. Expansion toward more domain-specific datasets and cutting-edge research from nanoparticles, photochemical catalysis, mass spectroscopy analysis, and protein design is also under consideration to support a broader range of research interests.

## CONCLUSION

AI4CHEM provides a discipline-aligned introduction to machine learning for experimental chemists, lowering barriers to entry through browser-based tools, chemically grounded examples, and active tutorial-driven instruction. Assessment data and student feedback indicate that the course builds both confidence and competence, preparing learners to read current AI-driven literature, communicate technical ideas, and begin integrating data-driven reasoning into their research. As AI continues to influence modern chemistry, introductory training tailored to experimental chemists will become increasingly important. Our experience suggests the open-access design increases the chance educators can adapt the curriculum for their own students, offering a practical path for expanding AI literacy across the synthetic chemistry community.

## ASSOCIATED CONTENT

**Supporting Information**. Supporting Information includes the full set of course goals for CHEM 5080: AI for Experimental Chemistry. This material is available free of charge at http://pubs.acs.org.
The complete course website, including all Google Colab tutorials and Jupyter Book materials, is available at: https://zhenglab.wustl.edu/chem5080.
All source code for the course website and tutorials is openly accessible at: https://github.com/zzhenglab/ai4chem.

## AUTHOR INFORMATION

Corresponding Author

**Zhiling Zheng** – Department of Chemistry, Washington University, St. Louis, Missouri 63130, United States

## ACKNOWLEDGMENT

Z.Z. thanks the Washington University Center for Teaching and Learning for its support through workshops and instructional development resources. Z.Z. is also grateful to Prof. Robert Wexler and Prof. Chenfeng Ke in the Department of Chemistry for their thoughtful discussions during the design of the course. The author also thanks the students enrolled in AI4CHEM for their engagement and feedback throughout the semester.

## REFERENCES

(1) Heid, E.; Greenman, K. P.; Chung, Y.; Li, S.-C.; Graff, D. E.; Vermeire, F. H.; Wu, H.; Green, W. H.; McGill, C. J. Chemprop: A Machine Learning Package for Chemical Property Prediction. *J. Chem. Inf. Model.* **2024**, *64* (1), 9–17. https://doi.org/10.1021/acs.jcim.3c01250.

(2) Pires, D. E. V.; Blundell, T. L.; Ascher, D. B. pkCSM: Predicting Small-Molecule Pharmacokinetic and Toxicity Properties Using Graph-Based Signatures. *J. Med. Chem.* **2015**, *58* (9), 4066–4072. https://doi.org/10.1021/acs.jmedchem.5b00104.

(3) Gilmer, J.; Schoenholz, S. S.; Riley, P. F.; Vinyals, O.; Dahl, G. E. Neural Message Passing for Quantum Chemistry. In *Proceedings of the 34th International Conference on Machine Learning - Volume 70*; ICML'17; JMLR.org: Sydney, NSW, Australia, 2017; pp 1263–1272.

(4) Yang, K.; Swanson, K.; Jin, W.; Coley, C.; Eiden, P.; Gao, H.; Guzman-Perez, A.; Hopper, T.; Kelley, B.; Mathea, M.; Palmer, A.; Settels, V.; Jaakkola, T.; Jensen, K.; Barzilay, R. Analyzing Learned Molecular Representations for Property Prediction. *J. Chem. Inf. Model.* **2019**, *59* (8), 3370–3388. https://doi.org/10.1021/acs.jcim.9b00237.

(5) Choi, J.; Kim, S.; Jung, Y. Synthesis-Aware Materials Redesign via Large Language Models. *J. Am. Chem. Soc.* **2025**, *147* (43), 39113–39122. https://doi.org/10.1021/jacs.5c07743.

(6) Zeni, C.; Pinsler, R.; Zügner, D.; Fowler, A.; Horton, M.; Fu, X.; Wang, Z.; Shysheya, A.; Crabbé, J.; Ueda, S.; Sordillo, R.; Sun, L.; Smith, J.; Nguyen, B.; Schulz, H.; Lewis, S.; Huang, C.-W.; Lu, Z.; Zhou, Y.; Yang, H.; Hao, H.; Li, J.; Yang, C.; Li, W.; Tomioka, R.; Xie, T. A Generative Model for Inorganic Materials Design. *Nature* **2025**, *639* (8055), 624–632. https://doi.org/10.1038/s41586-025-08628-5.

(7) Gómez-Bombarelli, R.; Wei, J. N.; Duvenaud, D.; Hernández-Lobato, J. M.; Sánchez-Lengeling, B.; Sheberla, D.; Aguilera-Iparraguirre, J.; Hirzel, T. D.; Adams, R. P.; Aspuru-Guzik, A. Automatic Chemical Design Using a Data-Driven Continuous Representation of Molecules. *ACS Cent. Sci.* **2018**, *4* (2), 268–276. https://doi.org/10.1021/acscentsci.7b00572.

(8) Schwaller, P.; Laino, T.; Gaudin, T.; Bolgar, P.; Hunter, C. A.; Bekas, C.; Lee, A. A. Molecular Transformer: A Model for Uncertainty-Calibrated Chemical Reaction Prediction. *ACS Cent. Sci.* **2019**, *5* (9), 1572–1583. https://doi.org/10.1021/acscentsci.9b00576.

(9) Torres, J. A. G.; Lau, S. H.; Anchuri, P.; Stevens, J. M.; Tabora, J. E.; Li, J.; Borovika, A.; Adams, R. P.; Doyle, A. G. A Multi-Objective Active Learning Platform and Web App for Reaction Optimization. *J. Am. Chem. Soc.* **2022**, *144* (43), 19999–20007. https://doi.org/10.1021/jacs.2c08592.

(10) Zheng, Z.; Zhang, O.; Nguyen, H. L.; Rampal, N.; Alawadhi, A. H.; Rong, Z.; Head-Gordon, T.; Borgs, C.; Chayes, J. T.; Yaghi, O. M. ChatGPT Research Group for Optimizing the Crystallinity of MOFs and COFs. *ACS Cent. Sci.* **2023**, *9* (11), 2161–2170. https://doi.org/10.1021/acscentsci.3c01087.

(11) Reid, J. P.; Proctor, R. S. J.; Sigman, M. S.; Phipps, R. J. Predictive Multivariate Linear Regression Analysis Guides Successful Catalytic Enantioselective Minisci Reactions of Diazines. *J. Am. Chem. Soc.* **2019**, *141* (48), 19178–19185. https://doi.org/10.1021/jacs.9b11658.

(12) Rinehart, N. I.; Saunthwal, R. K.; Wellauer, J.; Zahrt, A. F.; Schlemper, L.; Shved, A. S.; Bigler, R.; Fantasia, S.; Denmark, S. E. A Machine-Learning Tool to Predict Substrate-Adaptive Conditions for Pd-Catalyzed C–N Couplings. *Science* **2023**, *381* (6661), 965–972. https://doi.org/10.1126/science.adg2114.

(13) Coley, C. W.; Thomas, D. A.; Lummiss, J. A. M.; Jaworski, J. N.; Breen, C. P.; Schultz, V.; Hart, T.; Fishman, J. S.; Rogers, L.; Gao, H.; Hicklin, R. W.; Plehiers, P. P.; Byington, J.; Piotti, J. S.; Green, W. H.; Hart, A. J.; Jamison, T. F.; Jensen, K. F. A Robotic Platform for Flow Synthesis of Organic Compounds Informed by AI Planning. *Science* **2019**, *365* (6453), eaax1566. https://doi.org/10.1126/science.aax1566.

(14) Roch, L. M.; Häse, F.; Kreisbeck, C.; Tamayo-Mendoza, T.; Yunker, L. P. E.; Hein, J. E.; Aspuru-Guzik, A. ChemOS: Orchestrating Autonomous Experimentation. *Sci. Robot.* **2018**, *3* (19), eaat5559. https://doi.org/10.1126/scirobotics.aat5559.

(15) MacLeod, B. P.; Parlane, F. G. L.; Morrissey, T. D.; Häse, F.; Roch, L. M.; Dettelbach, K. E.; Moreira, R.; Yunker, L. P. E.; Rooney, M. B.; Deeth, J. R.; Lai, V.; Ng, G. J.; Situ, H.; Zhang, R. H.; Elliott, M. S.; Haley, T. H.; Dvorak, D. J.; Aspuru-Guzik, A.; Hein, J. E.; Berlinguette, C. P. Self-Driving Laboratory for Accelerated Discovery of Thin-Film Materials. *Sci. Adv.* **2020**, *6* (20), eaaz8867. https://doi.org/10.1126/sciadv.aaz8867.

(16) Koscher, B. A.; Canty, R. B.; McDonald, M. A.; Greenman, K. P.; McGill, C. J.; Bilodeau, C. L.; Jin, W.; Wu, H.; Vermeire, F. H.; Jin, B.; Hart, T.; Kulesza, T.; Li, S.-C.; Jaakkola, T. S.; Barzilay, R.; Gómez-Bombarelli, R.; Green, W. H.; Jensen, K. F. Autonomous, Multiproperty-Driven Molecular Discovery: From Predictions to Measurements and Back. *Science* **2023**, *382* (6677), eadi1407. https://doi.org/10.1126/science.adi1407.

(17) Steiner, S.; Wolf, J.; Glatzel, S.; Andreou, A.; Granda, J. M.; Keenan, G.; Hinkley, T.; Aragon-Camarasa, G.; Kitson, P. J.; Angelone, D.; Cronin, L. Organic Synthesis in a Modular Robotic System Driven by a Chemical Programming Language. *Science* **2019**, *363* (6423), eaav2211. https://doi.org/10.1126/science.aav2211.

(18) Baum, Z. J.; Yu, X.; Ayala, P. Y.; Zhao, Y.; Watkins, S. P.; Zhou, Q. Artificial Intelligence in Chemistry: Current Trends and Future Directions. *J. Chem. Inf. Model.* **2021**, *61* (7), 3197–3212. https://doi.org/10.1021/acs.jcim.1c00619.

(19) Wang, H.; Fu, T.; Du, Y.; Gao, W.; Huang, K.; Liu, Z.; Chandak, P.; Liu, S.; Van Katwyk, P.; Deac, A.; Anandkumar, A.; Bergen, K.; Gomes, C. P.; Ho, S.; Kohli, P.; Lasenby, J.; Leskovec, J.; Liu, T.-Y.; Manrai, A.; Marks, D.; Ramsundar, B.; Song, L.; Sun, J.; Tang, J.; Veličković, P.; Welling, M.; Zhang, L.; Coley, C. W.; Bengio, Y.; Zitnik, M. Scientific Discovery in the Age of Artificial Intelligence. *Nature* **2023**, *620* (7972), 47–60. https://doi.org/10.1038/s41586-023-06221-2.

(20) de Almeida, A. F.; Moreira, R.; Rodrigues, T. Synthetic Organic Chemistry Driven by Artificial Intelligence. *Nat. Rev. Chem.* **2019**, *3* (10), 589–604. https://doi.org/10.1038/s41570-019-0124-0.

(21) Zheng, Z.; Rampal, N.; Inizan, T. J.; Borgs, C.; Chayes, J. T.; Yaghi, O. M. Large Language Models for Reticular Chemistry. *Nat. Rev. Mater.* **2025**, *10* (5), 369–381. https://doi.org/10.1038/s41578-025-00772-8.

(22) Kim, S.-Y.; Jeon, I.; Kang, S.-J. Integrating Data Science and Machine Learning to Chemistry Education: Predicting Classification and Boiling Point of Compounds. *J. Chem. Educ.* **2024**, *101* (4), 1771–1776. https://doi.org/10.1021/acs.jchemed.3c01040.

(23) Iyamuremye, A.; Niyonzima, F. N.; Mukiza, J.; Twagilimana, I.; Nyirahabimana, P.; Nsengimana, T.; Habiyaremye, J. D.; Habimana, O.; Nsabayezu, E. Utilization of Artificial Intelligence and Machine Learning in Chemistry Education: A Critical Review. *Discov. Educ.* **2024**, *3* (1), 95. https://doi.org/10.1007/s44217-024-00197-5.

(24) Yuriev, E.; Wink, D. J.; Holme, T. A. The Dawn of Generative Artificial Intelligence in Chemistry Education. *J. Chem. Educ.* **2024**, *101* (8), 2957–2959. https://doi.org/10.1021/acs.jchemed.4c00836.

(25) Lafuente, D.; Cohen, B.; Fiorini, G.; García, A. A.; Bringas, M.; Morzan, E.; Onna, D. A Gentle Introduction to Machine Learning for Chemists: An Undergraduate Workshop Using Python Notebooks for Visualization, Data Processing, Analysis, and Modeling. *J. Chem. Educ.* **2021**, *98* (9), 2892–2898. https://doi.org/10.1021/acs.jchemed.1c00142.

(26) Xue, M.; Liu, B.; Huang, X. Machine Learning in Chemistry: A Data Centred, Hands-on Introductory Machine Learning Course for Undergraduate Students. Chemistry October 7, 2025. https://doi.org/10.26434/chemrxiv-2025-9zldf.

(27) Fuchs, W.; McDonald, A. R.; Gautam, A.; Kazerouni, A. M. Recommendations for Improving End-User Programming Education: A Case Study with Undergraduate Chemistry Students. *J. Chem. Educ.* **2024**, *101* (8), 3085–3096. https://doi.org/10.1021/acs.jchemed.4c00219.

(28) Caccavale, F.; Gargalo, C. L.; Gernaey, K. V.; Krühne, U. SPyCE: A Structured and Tailored Series of Python Courses for (Bio)Chemical Engineers. *Educ. Chem. Eng.* **2023**, *45*, 90–103. https://doi.org/10.1016/j.ece.2023.08.003.

(29) Kim, S.; Bucholtz, E. C.; Briney, K.; Cornell, A. P.; Cuadros, J.; Fulfer, K. D.; Gupta, T.; Hepler-Smith, E.; Johnston, D. H.; Lang, A. S. I. D.; Larsen, D.; Li, Y.; McEwen, L. R.; Morsch, L. A.; Muzyka, J. L.; Belford, R. E. Teaching Cheminformatics through a Collaborative Intercollegiate Online Chemistry Course (OLCC). *J. Chem. Educ.* **2021**, *98* (2), 416–425. https://doi.org/10.1021/acs.jchemed.0c01035.

(30) Remington, J. M.; Ferrell, J. B.; Zorman, M.; Petrucci, A.; Schneebeli, S. T.; Li, J. Machine Learning in a Molecular Modeling Course for Chemistry, Biochemistry, and Biophysics Students. **2020**. https://doi.org/10.35459/tbp.2019.000140.

(31) Varnek, A.; Marcou, G.; Horvath, D. Higher Education in Chemoinformatics: Achievements and Challenges. *J. Cheminformatics* **2025**, *17* (1), 89. https://doi.org/10.1186/s13321-025-01036-x.

(32) Hughes, D. J.; Perry, S. C. Modular Integration of Python Programming in Undergraduate Physical Chemistry Experiments. *J. Chem. Educ.* **2025**, *102* (9), 4005–4016. https://doi.org/10.1021/acs.jchemed.5c00677.

(33) Blonder, R.; Feldman-Maggor, Y. AI for Chemistry Teaching: Responsible AI and Ethical Considerations. *Chem. Teach. Int.* **2024**, *6* (4), 385–395. https://doi.org/10.1515/cti-2024-0014.

(34) Berber, S.; Brückner, M.; Maurer, N.; Huwer, J. Artificial Intelligence in Chemistry Research—Implications for Teaching and Learning. *J. Chem. Educ.* **2025**, *102* (4), 1445–1456. https://doi.org/10.1021/acs.jchemed.4c01033.

(35) Zheng, Z. The Future of Reviews Writing in the AI Era. *Nat. Rev. Chem.* **2025**, 1–2. https://doi.org/10.1038/s41570-025-00738-y.

(36) Neumann, S. E.; Neumann, K.; Zheng, Z.; Hanikel, N.; Tsao, J.; Yaghi, O. M. Harvesting Water in the Classroom. *J. Chem. Educ.* **2023**, *100* (11), 4482–4487. https://doi.org/10.1021/acs.jchemed.3c00690.

(37) Tu, Z.; Choure, S. J.; Fong, M. H.; Roh, J.; Levin, I.; Yu, K.; Joung, J. F.; Morgan, N.; Li, S.-C.; Sun, X.; Lin, H.; Murnin, M.; Liles, J. P.; Struble, T. J.; Fortunato, M. E.; Liu, M.; Green, W. H.; Jensen, K. F.; Coley, C. W. ASKCOS: Open-Source, Data-Driven Synthesis Planning. *Acc. Chem. Res.* **2025**, *58* (11), 1764–1775. https://doi.org/10.1021/acs.accounts.5c00155.

(38) Abramson, J.; Adler, J.; Dunger, J.; Evans, R.; Green, T.; Pritzel, A.; Ronneberger, O.; Willmore, L.; Ballard, A. J.; Bambrick, J.; Bodenstein, S. W.; Evans, D. A.; Hung, C.-C.; O'Neill, M.; Reiman, D.; Tunyasuvunakool, K.; Wu, Z.; Žemgulytė, A.; Arvaniti, E.; Beattie, C.; Bertolli, O.; Bridgland, A.; Cherepanov, A.; Congreve, M.; Cowen-Rivers, A. I.; Cowie, A.; Figurnov, M.; Fuchs, F. B.; Gladman, H.; Jain, R.; Khan, Y. A.; Low, C. M. R.; Perlin, K.; Potapenko, A.; Savy, P.; Singh, S.; Stecula, A.; Thillaisundaram, A.; Tong, C.; Yakneen, S.; Zhong, E. D.; Zielinski, M.; Žídek, A.; Bapst, V.; Kohli, P.; Jaderberg, M.; Hassabis, D.; Jumper, J. M. Accurate Structure Prediction of Biomolecular Interactions with AlphaFold 3. *Nature* **2024**, *630* (8016), 493–500. https://doi.org/10.1038/s41586-024-07487-w.

(39) Gao, W.; Coley, C. W. The Synthesizability of Molecules Proposed by Generative Models. *J. Chem. Inf. Model.* **2020**, *60* (12), 5714–5723. https://doi.org/10.1021/acs.jcim.0c00174.

(40) Zheng, Z.; Zhang, O.; Borgs, C.; Chayes, J. T.; Yaghi, O. M. ChatGPT Chemistry Assistant for Text Mining and the Prediction of MOF Synthesis. *J. Am. Chem. Soc.* **2023**, *145* (32), 18048–18062. https://doi.org/10.1021/jacs.3c05819.

(41) Weininger, D. SMILES, a Chemical Language and Information System. 1. Introduction to Methodology and Encoding Rules. *J. Chem. Inf. Comput. Sci.* **1988**, *28* (1), 31–36. https://doi.org/10.1021/ci00057a005.

(42) Pedregosa, F.; Varoquaux, G.; Gramfort, A.; Michel, V.; Thirion, B.; Grisel, O.; Blondel, M.; Prettenhofer, P.; Weiss, R.; Dubourg, V.; Vanderplas, J.; Passos, A.; Cournapeau, D.; Brucher, M.; Perrot, M.; Duchesnay, É. Scikit-Learn: Machine Learning in Python. *J Mach Learn Res* **2011**, *12* (null), 2825–2830.

(43) Kim, S.; Thiessen, P. A.; Bolton, E. E.; Chen, J.; Fu, G.; Gindulyte, A.; Han, L.; He, J.; He, S.; Shoemaker, B. A.; Wang, J.; Yu, B.; Zhang, J.; Bryant, S. H. PubChem Substance and Compound Databases. *Nucleic Acids Res.* **2016**, *44* (D1), D1202-1213. https://doi.org/10.1093/nar/gkv951.

(44) Paszke, A.; Gross, S.; Massa, F.; Lerer, A.; Bradbury, J.; Chanan, G.; Killeen, T.; Lin, Z.; Gimelshein, N.; Antiga, L.; Desmaison, A.; Kopf, A.; Yang, E.; DeVito, Z.; Raison, M.; Tejani, A.; Chilamkurthy, S.; Steiner, B.; Fang, L.; Bai, J.; Chintala, S. PyTorch: An Imperative Style, High-Performance Deep Learning Library. In *Advances in Neural Information Processing Systems*; Curran Associates, Inc., 2019; Vol. 32.

(45) Zheng, Z.; Florit, F.; Jin, B.; Wu, H.; Li, S.-C.; Nandiwale, K. Y.; Salazar, C. A.; Mustakis, J. G.; Green, W. H.; Jensen, K. F. Integrating Machine Learning and Large Language Models to Advance Exploration of Electrochemical Reactions. *Angew. Chem. Int. Ed.* **2024**, *63* (63), e202418074.

(46) Ramakrishnan, R.; Dral, P. O.; Rupp, M.; von Lilienfeld, O. A. Quantum Chemistry Structures and Properties of 134 Kilo Molecules. *Sci. Data* **2014**, *1* (1), 140022. https://doi.org/10.1038/sdata.2014.22.

(47) Lecun, Y.; Bottou, L.; Bengio, Y.; Haffner, P. Gradient-Based Learning Applied to Document Recognition. *Proc. IEEE* **1998**, *86* (11), 2278–2324. https://doi.org/10.1109/5.726791.

(48) Chong, K. X.; Alsabia, Q. A.; Ye, Z.; McDaniel, A.; Baumgardner, D.; Xiao, D.; Sun, S. Controlling Metal–Organic Framework Crystallization via Computer Vision and Robotic Handling. *J. Mater. Chem. A* **2025**, *13* (33), 27279–27289. https://doi.org/10.1039/D5TA03199K.

(49) Vaswani, A.; Shazeer, N.; Parmar, N.; Uszkoreit, J.; Jones, L.; Gomez, A. N.; Kaiser, Ł. ukasz; Polosukhin, I. Attention Is All You Need. In *Advances in Neural Information Processing Systems*; Curran Associates, Inc., 2017; Vol. 30.

(50) Zheng, Z.; Rong, Z.; Rampal, N.; Borgs, C.; Chayes, J. T.; Yaghi, O. M. A GPT-4 Reticular Chemist for Guiding MOF Discovery. *Angew Chem Int Ed* **2023**, *62* (46), e202311983. https://doi.org/10.1002/anie.202311983.

(51) M. Bran, A.; Cox, S.; Schilter, O.; Baldassari, C.; White, A. D.; Schwaller, P. Augmenting Large Language Models with Chemistry Tools. *Nat. Mach. Intell.* **2024**, *6* (5), 525–535. https://doi.org/10.1038/s42256-024-00832-8.

(52) Li, A.; Jabri, A.; Joulin, A.; Van Der Maaten, L. Learning Visual N-Grams from Web Data. In *2017 IEEE International Conference on Computer Vision (ICCV)*; IEEE: Venice, 2017; pp 4193–4202. https://doi.org/10.1109/ICCV.2017.449.

(53) Radford, A.; Kim, J. W.; Hallacy, C.; Ramesh, A.; Goh, G.; Agarwal, S.; Sastry, G.; Askell, A.; Mishkin, P.; Clark, J.; Krueger, G.; Sutskever, I. Learning Transferable Visual Models From Natural Language Supervision. arXiv February 26, 2021. https://doi.org/10.48550/arXiv.2103.00020.

(54) Zheng, Z.; He, Z.; Khattab, O.; Rampal, N.; Zaharia, M. A.; Borgs, C.; Chayes, J. T.; Yaghi, O. M. Image and Data Mining in Reticular Chemistry Powered by GPT-4V. *Digit. Discov.* **2024**, *3* (3), 491–501. https://doi.org/10.1039/D3DD00239J.

(55) Dagdelen, J.; Dunn, A.; Lee, S.; Walker, N.; Rosen, A. S.; Ceder, G.; Persson, K. A.; Jain, A. Structured Information Extraction from Scientific Text with Large Language Models. *Nat. Commun.* **2024**, *15* (1), 1418. https://doi.org/10.1038/s41467-024-45563-x.

(56) Shields, B. J.; Stevens, J.; Li, J.; Parasram, M.; Damani, F.; Alvarado, J. I. M.; Janey, J. M.; Adams, R. P.; Doyle, A. G. Bayesian Reaction Optimization as a Tool for Chemical Synthesis. *Nature* **2021**, *590* (7844), 89–96. https://doi.org/10.1038/s41586-021-03213-y.

(57) Park, H.; Majumdar, S.; Zhang, X.; Kim, J.; Smit, B. Inverse Design of Metal–Organic Frameworks for Direct Air Capture of $CO_2$ via Deep Reinforcement Learning. *Digit. Discov.* **2024**, *3* (4), 728–741. https://doi.org/10.1039/D4DD00010B.

(58) Wang, J. Y.; Stevens, J. M.; Kariofillis, S. K.; Tom, M.-J.; Golden, D. L.; Li, J.; Tabora, J. E.; Parasram, M.; Shields, B. J.; Primer, D. N.; Hao, B.; Del Valle, D.; DiSomma, S.; Furman, A.; Zipp, G. G.; Melnikov, S.; Paulson, J.; Doyle, A. G. Identifying General Reaction Conditions by Bandit Optimization. *Nature* **2024**, *626* (8001), 1025–1033. https://doi.org/10.1038/s41586-024-07021-y.

(59) Park, H.; Kang, Y.; Choe, W.; Kim, J. Mining Insights on Metal–Organic Framework Synthesis from Scientific Literature Texts. *J. Chem. Inf. Model.* **2022**, *62* (5), 1190–1198. https://doi.org/10.1021/acs.jcim.1c01297.

(60) Zheng, Z.; Alawadhi, A. H.; Chheda, S.; Neumann, S. E.; Rampal, N.; Liu, S.; Nguyen, H. L.; Lin, Y.; Rong, Z.; Siepmann, J. I.; Gagliardi, L.; Anandkumar, A.; Borgs, C.; Chayes, J. T.; Yaghi, O. M. Shaping the Water-Harvesting Behavior of Metal–Organic Frameworks Aided by Fine-Tuned GPT Models. *J. Am. Chem. Soc.* **2023**, *145* (51), 28284–28295. https://doi.org/10.1021/jacs.3c12086.

(61) Zheng, Z.; Alawadhi, A. H.; Yaghi, O. M. Green Synthesis and Scale-Up of MOFs for Water Harvesting from Air. *Mol. Front. J.* **2023**, *07* (01n02), 20–39. https://doi.org/10.1142/S2529732523400011.

(62) Ruff, E. F.; Zemke, J. M. O. Discussing the Ethics of Professional AI Use in Undergraduate Chemistry Courses. *J. Chem. Educ.* **2025**, *102* (4), 1457–1464. https://doi.org/10.1021/acs.jchemed.4c01038.

# Course Goals for CHEM 5080: AI for Experimental Chemistry

By the end of the course, students will be able to:

- Identify and select appropriate machine learning approaches for a wide range of chemical data analysis and prediction tasks.
- Communicate ML concepts in chemistry clearly through figures, visualizations, written reports, and oral presentations.
- Interpret common cheminformatics workflows and reuse ready-to-run code for experimental planning and analysis.
- Apply classification and regression methods to predict physical and chemical properties, reaction outcomes, and spectroscopic signals.
- Implement and evaluate neural-network–based models for molecular property prediction and design.
- Analyze imaging data such as microscope and electron microscopy images using computer vision techniques.
- Use generative and optimization methods to propose new experiments or molecules.
- Prompt or fine-tune transformer-based language models (e.g., GPT-5) to summarize literature, plan experiments, troubleshoot code, and support self-driving lab workflows.

| Class Day (80-minute classes, 2 days/week) | Topic | Learning Objectives |
|---|---|---|
| 1 | Introduction to Course and Python Primer | 1. Describe the goals, structure, assessments, and expectations of CHEM 5080.<br>2. Navigate the Jupyter Book, Colab notebooks, and course resources.<br>3. Run code and Markdown cells in a notebook and switch between the two modes.<br>4. Use Python as a calculator for basic chemical math (e.g., moles, molar mass).<br>5. Store values in variables, create simple lists and dictionaries, and access their elements. |
| 2 | Pandas and Plotting for Chemical Data | 1. Explain what *pandas* is, define *Series* and *DataFrame*, and use standard naming conventions.<br>2. Read CSV files into a *DataFrame*, inspect column types, and perform basic cleaning (sorting, filtering, handling missing values).<br>3. Select, filter, group, and summarize data from chemical datasets using pandas.<br>4. Create line, scatter, bar, histogram, box, violin, and heatmap plots with Matplotlib.<br>5. Combine pandas and plotting to explore real chemical data (e.g., Beer–Lambert–law examples) and save publication-quality figures. |
| 3 | SMILES and RDKit: Machine-Readable Molecules | 1. Interpret SMILES strings in terms of atoms, bonds, branches, rings, aromaticity, charges, and simple stereochemistry.<br>2. Use RDKit to parse SMILES, draw molecular structures, add hydrogens, and compute basic molecular properties.<br>3. Perform small structure edits in RDKit (e.g., atom substitution, neutralizing groups, adding a methyl group).<br>4. Connect to PubChem to retrieve SMILES and related information, then round-trip between text, RDKit objects, and files. |
| 4 | Chemical Structure Identifiers and Web Services | 1. Describe PubChem's APIs as chemical data services and explain typical use cases.<br>2. Construct URLs that return JSON, text, or images for given identifiers (name, SMILES, CAS, CID).<br>3. Resolve chemical names, SMILES, and CAS numbers to PubChem CIDs and retrieve IUPAC names, SMILES, InChIKeys, and selected properties.<br>4. Use the NCI Chemical Identifier Resolver (CIR) as a second query path and compare its responses to PubChem.<br>5. Write small helper functions with basic error handling and fallbacks to automate identifier resolution for a list of ligands. |

| 5 | Regression and Classification with Chemical Data | 1. Distinguish between regression and classification problems by examining the type of target variable. |
| --- | --- | --- |
| | | 2. Load small chemistry datasets containing SMILES and simple descriptors or text features. |
| | | 3. Create train, validation, and test splits and describe the role of each split in model development. |
| | | 4. Fit basic regression model using linear regression and logistic regression. |
| | | 5. Compute and interpret standard metrics including RMSE, MAE, $R^2$, accuracy, precision, recall, F1, and ROC-AUC to compare models. |
| 6 | Cross-Validation, Model Selection, and Feature Importance | 1. Use K-fold cross-validation to obtain fairer performance estimates than a single train/test split. |
| | | 2. Explain the role of hyperparameters and tune them with tools such as *GridSearchCV*. |
| | | 3. Perform basic exploratory data analysis by plotting descriptor distributions, pair plots, and correlations. |
| | | 4. Apply cross-validation to compare models and hyperparameter settings, then choose a final model. |
| | | 5. Interpret feature importance measures to explain model predictions on chemical properties. |
| 7 | Decision Trees and Random Forests | 6. Describe the intuition behind decision trees for both regression and classification problems. |
| | | 7. Interpret Gini impurity, entropy, and mean squared error as criteria for splitting nodes. |
| | | 8. Grow and visualize a decision tree, examining nodes, depth, and leaf counts. |
| | | 9. Control overfitting using hyperparameters. |
| | | 10. Train random forest models for toxicity or property prediction and compare their performance to single trees. |
| | | 11. Use tree-based feature importance and permutation importance to identify key molecular descriptors. |
| 8 | Introduction to Neural Networks | • Explain the components of a multilayer perceptron (MLP). |
| | | • Build a small MLP for a toy dataset, then extend it to chemical tasks such as solubility or toxicity prediction. |
| | | • Train *MLPRegressor* and *MLPClassifier* models step by step and monitor training and validation curves. |
| | | • Recognize signs of overfitting and describe strategies such as regularization, early stopping, and data scaling. |
| | | • Compare neural-network performance with simpler baselines (e.g., linear models) on the same chemical dataset. |
| 9 | Graph Neural Networks for Molecules | 1. Represent molecules as graphs with atoms as nodes, bonds as edges, and appropriate node and edge features. |
| | | 2. Build a basic MLP in *PyTorch* and use it as a stepping stone to graph neural networks (GNNs). |
| | | 3. Explain message passing and neighborhood aggregation in message-passing neural networks (MPNNs). |
| | | 4. Implement a tiny GNN in *PyTorch* to predict properties on toy molecular graphs. |
| | | 5. Prepare molecular graphs from SMILES and run a simple GNN model, comparing its performance to descriptor-based models. |
| 10 | Property and Reaction Prediction with Graph Neural Networks | 1. Set up message-passing neural networks (D-MPNNs) for both regression and classification tasks on a reaction dataset (e.g., C–H oxidation). |
| | | 2. Train single-task *Chemprop* models for properties such as solubility, pKa, melting point, and toxicity. |
| | | 3. Train a reactivity classifier and an atom-level selectivity predictor for reaction outcomes. |
| | | 4. Interpret *Chemprop* models using Shapley values (SHAP) at the feature and node levels. |
| 11 | Dimension Reduction and Visualization | 1. Differentiate supervised from unsupervised learning in a chemistry context. |
| | | 2. Explain the intuition and basic mathematics of Principal Component Analysis (PCA) and interpret loadings, scores, and explained variance. |
| | | 3. Use t-SNE and UMAP to embed high-dimensional chemical features into 2D for visualization. |
| | | 4. Compare descriptor-based and fingerprint-based representations in low-dimensional plots. |
| | | 5. Use distance metrics and clustering outputs to explore structure–property relationships in a reaction dataset. |

| 12 | **Clustering and Self-Supervised Workflows** | 1. Build clustering pipelines that include feature selection, scaling, clustering, and visualization.<br>2. Select suitable distance metrics for descriptors versus fingerprints and justify these choices.<br>3. Use K-means clustering and evaluate candidate values of k using elbow and silhouette analyses.<br>4. Explore alternative clustering methods such as agglomerative clustering and DBSCAN and compare their behavior.<br>5. Interpret clustering results in terms of chemical similarity, reactivity, or experimental outcomes. |
|---|---|---|
| 13 | **De Novo Molecule Generation with Variational Autoencoders** | 1. Connect unsupervised learning concepts such as reconstruction and latent space to molecular generation tasks.<br>2. Explain encoder and decoder roles in a variational autoencoder (VAE) and why VAEs are useful for sampling.<br>3. Train a small SMILES-based VAE model on a molecular dataset.<br>4. Inspect latent-space organization and perform simple sampling or interpolation to generate new molecules.<br>5. Discuss the strengths and limitations of VAE-based generative models for molecular design. |
| 14 | **Bayesian Optimization for Synthesis Conditions** | 1. Describe the motivation for Bayesian optimization (BO) in expensive experimental settings.<br>2. Define key components of BO: prior, surrogate model (GP, RF, small NN), and acquisition function (EI, UCB, PI, greedy).<br>3. Implement a basic BO loop: fit surrogate, compute acquisition, pick the next point, update data, and repeat.<br>4. Visualize BO behavior in 1D or low-dimensional examples to build intuition about exploration–exploitation trade-offs.<br>5. Apply BO to a toy Suzuki coupling dataset to optimize yield over temperature, time, and concentration. |
| 15 | **Multi-Objective Bayesian Optimization** | 1. Extend single-objective BO concepts to multi-objective problems common in chemistry (e.g., yield, purity, and cost).<br>2. Define Pareto dominance, Pareto front, scalarization, hypervolume, and expected hypervolume improvement.<br>3. Engineer features and targets for a multi-objective metal-organic framework (MOF) synthesis dataset.<br>4. Build simple surrogate models for each objective and use them within a multi-objective BO loop.<br>5. Analyze and visualize Pareto fronts to support decision-making in multi-criteria experimental design. |
| 16 | **Reinforcement Learning and Bandits for Experiment Design** | 1. Define agent, environment, state, action, reward, trajectory, policy, and value in reinforcement learning (RL).<br>2. Implement tabular Q-learning in a simple gridworld with a chemistry-inspired reward structure.<br>3. Compare exploration strategies such as ε-greedy, optimistic initialization, UCB, and Thompson sampling in bandit problems.<br>4. Frame a chemistry example (e.g., MOF synthesis) as a multi-armed bandit and simulate different agents.<br>5. Explain how RL and bandit methods can inform closed-loop experiment selection. |
| 17 | **Positive–Unlabeled (PU) Learning** | 1. Define semi-supervised learning and distinguish between LU (labeled + unlabeled) and PU (positive + unlabeled) settings.<br>2. Explain standard assumptions used in PU learning and when they are reasonable in chemical datasets.<br>3. Construct a simple PU workflow for a chemistry example where failures are unlabeled or rarely reported.<br>4. Estimate class priors and convert scores from an intermediate classifier into PU probabilities.<br>5. Propose evaluation strategies when true negatives are unavailable or scarce. |
| 18 | **Contrastive Learning and Data Augmentation** | 6. Explain the key ideas of contrastive learning and its relation to representation learning.<br>7. Construct positive and negative pairs using easy data augmentations that mimic realistic lab variability.<br>8. Derive and implement a small-scale InfoNCE loss with arrays before applying it to models.<br>9. Design and justify augmentations for spectra, reaction conditions, or other chemical measurements. |
| | | 10. Evaluate learned embeddings using linear probes and retrieval tasks based on cosine similarity. |
| 19 | **Transformers for Chemistry** | 1. Define tokens, embeddings, positional encodings, attention, and multi-head attention in transformer models.
2. Convert SMILES strings into token sequences suitable for input to a transformer encoder.
3. Implement a minimal transformer encoder block and train it on a small chemistry task.
4. Visualize attention patterns and discuss what the model might be "looking at" in a SMILES string.
5. Compare transformer-based models to earlier MLP and GNN architectures on similar tasks. |
| 20 | **Large Language Models** | 1. Use the API to build a simple chemistry-focused chatbot.
2. Conduct both single-turn and multi-turn conversations while tracking context and message history.
3. Compare several LLM models on the same prompt in terms of latency, cost, and quality of responses.
4. Describe and monitor hallucinations in LLM outputs and discuss mitigation strategies (including retrieval-augmented generation). |
| 21 | **Computer Vision** | 1. Build and train a compact convolutional neural network (CNN) in *PyTorch* on a benchmark dataset such as MNIST.
2. Track learning curves, compute key metrics, and inspect example predictions.
3. Visualize learned filters and intermediate feature maps to interpret what the network has captured.
4. Connect CNN concepts to potential chemical applications such as micrograph analysis or crystal classification. |
| 22 | **Vision–Language Models** | 1. Explain the core idea of CLIP method: aligning image and text embeddings using a contrastive loss.
2. Build a simple zero-shot classifier for crystal or materials images using frozen CLIP embeddings.
3. Train a small linear probe on top of frozen CLIP embeddings and compare performance to zero-shot predictions.
4. Visualize the embedding structure of images and captions using PCA or t-SNE.
5. Start a basic vision-chat workflow that uses an LLM to reason over CLIP-style image and text representations. |
| 23 | **Prompt Engineering and Function Calling** | 1. Explain zero-shot, one-shot, and few-shot prompting and recognize when each style is appropriate in a chemistry setting.
2. Write structured prompts that request JSON or other machine-readable outputs for downstream analysis.
3. Design prompts that incorporate domain constraints and safety checks for chemical tasks.
4. Use function calling (or tool calling) from an LLM to run external utilities (e.g., unit conversion, RDKit calculations) and combine results in a final response.
5. Reflect on limitations, error modes, and ethical considerations when using LLMs to propose or modify experiments. |
| 24 | **Literature Data Mining with LLMs** | 1. Use an LLM to classify literature abstracts and prioritize papers for data mining.
2. Apply prompt-engineering strategies and domain heuristics to improve the consistency of abstract classifications.
3. Extract structured synthesis or reaction data from full-text PDFs using reasoning-augmented prompts and JSON schemas.
4. Implement basic vision-based figure recognition to locate and interpret plots, diagrams, or reaction schemes in papers.
5. Combine text- and image-based signals to build a small literature mining pipeline for a chemistry topic of interest. |
| 25 | **Toward Self-Driving Chemistry Labs** | 1. Understand how closed-loop workflows connect experimental design, automation, measurement, and ML models.
2. Explain how literature mining, property prediction, and optimization models can feed into self-driving lab systems.
3. Propose a simple self-driving-lab scenario for a chosen chemistry problem, identifying sensors, actuators, models, and decision logic.
4. Discuss practical challenges in deploying self-driving labs, including data quality, safety constraints, and human oversight. |

| | | 5. Reflect on the future roles of chemists in AI-enabled laboratories and identify skills that students can continue to develop beyond the course. |
|---|---|---|